\documentclass[letterpaper, 10 pt, conference]{ieeeconf}  

\IEEEoverridecommandlockouts                              

\usepackage{times}
\usepackage{url}
\usepackage{graphicx}
\usepackage{amsmath}
\usepackage{booktabs}
\usepackage{algorithm}
\usepackage{algorithmic}
\usepackage{multirow}
\usepackage{xcolor}
\graphicspath{ {./images/} }

\title{\LARGE \bf
Tidying Deep Saliency Prediction Architectures
}

\author{Navyasri Reddy$^{*1}$, Samyak Jain$^{*1}$, Pradeep Yarlagadda$^{*1}$ and Vineet Gandhi$^{1}$ 
\thanks{* Equal Contribution}
\thanks{$^{1}$ CVIT, KCIS, International Institute of Information Technology Hyderabad, India 
        {\tt\small navyasri.reddy@research.iiit.ac.in}}%
}

\begin{document}

\maketitle
\thispagestyle{empty}
\pagestyle{empty}


\begin{figure*}[h]
\includegraphics[width=\textwidth]{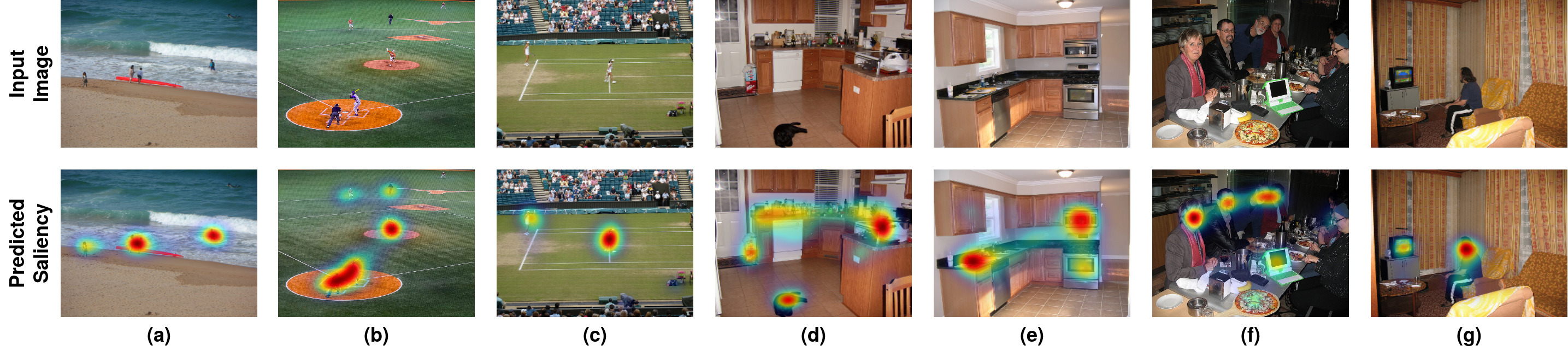}
\caption{Example results of our approach on images from Salicon dataset. Saliency models can play key role in application like (a) drone surveillance , (b,c) robotics cameras for sports , (d,e) indoor navigation and (f,g) social interactions }.
\label{fig:motivation}
\end{figure*}

\begin{abstract}

Learning computational models for visual attention (saliency estimation) is an effort to inch machines/robots closer to human visual cognitive abilities. Data-driven efforts have dominated the landscape since the introduction of deep neural network architectures. In deep learning research, the choices in architecture design are often empirical and frequently lead to more complex models than necessary. The complexity, in turn, hinders the application requirements. In this paper, we identify four key components of saliency models, i.e., input features,  multi-level integration, readout architecture, and loss functions. We review the existing state of the art models on these four components and propose novel and simpler alternatives. As a result, we propose two novel end-to-end architectures called SimpleNet and MDNSal, which are neater, minimal, more interpretable and achieve state of the art performance on public saliency benchmarks. SimpleNet is an optimized encoder-decoder architecture and brings notable performance gains on the SALICON dataset (the largest saliency benchmark). MDNSal is a parametric model that directly predicts parameters of a GMM distribution and is aimed to bring more interpretability to the prediction maps. The proposed saliency models can be inferred at 25fps, making them suitable for real-time applications. Code and pre-trained models are available at  https://github.com/samyak0210/saliency.

\end{abstract}

\section{Introduction}

Predicting the salient regions in a scene is a fundamental ability, which empowers primates to rapidly analyze/interpret the complex surroundings by locating and devoting the focus only on sub-regions of interest~\cite{koch1987shifts}. The work by~\cite{itti1998model} triggered early interest in the computational modeling of visual saliency from images, i.e., identifying areas that are salient in a scene. Since then, a large variety of saliency detection models have been proposed and find usages in a wide range of applications involving machine vision. Many recent works show that availability of saliency maps enhance cognitive abilities of robots and helps improving performance in variety of tasks including human-robot interaction~\cite{staudte2009visual}; identification, and recognition of objects~\cite{schauerte2010saliency}; scene classification~\cite{borji2011scene}; detecting and tracking regions of interest~\cite{frintrop2009most}; proposal refinement~\cite{chen2017saliency} and visual search in unknown environments (allowing search of regions with higher importance first)~\cite{rasouli2019attention}. Our work is application agnostic and focuses on improving the general saliency prediction and can cater to large variety of applications in robotic vision. Some example results from our SimpleNet model are illustrated in Figure~\ref{fig:motivation}.

Formally, computational saliency models predict the probability distribution of the location of the eye fixations over the image, i.e., the saliency map. Where human observers look in images is often used as a ground truth estimate of image saliency.  The predictions are evaluated using a variety of metrics, which are broadly classified as location-based or distribution-based~\cite{bylinskii2018different}. The location-based metrics measure the accuracy of saliency models at predicting discrete fixation locations. Distribution based metrics compute the difference/similarity between predicted and ground truth distributions (assuming that the ground truth fixation locations are sampled from an underlying probability distribution). 

The last few years have seen tremendous advancements in the field, mainly due to the application of Deep Neural Network architectures for the task and the availability of large scale datasets~\cite{jiang2015salicon}. Recent works have analyzed the saliency estimation models over different evaluation metrics to add interpretability to saliency scores~\cite{bylinskii2018different}. Interestingly, the interpretability of saliency model architectures has not been systematically explored. To this end, we propose a componential analysis which can be used to compare a model from another; reduce redundancies in the model without compromising the performance and can help customizations based on application requirements. 

We identify four key components in saliency models. First is the input features i.e., to directly send the image to the saliency models or employ transfer learning using pre-trained networks. The second component is the multi-level integration. It is understood that multiscale features (at different spatial and semantic hierarchy) capture a broad spectrum of stimuli, and the combination improves model performance. This aspect concerns how the hierarchy in imbibed in the model. The third aspect is the readout architecture, which concerns the form of output i.e., to directly predict a saliency map or to predict parameters of an assumed underlying distribution. The fourth aspect is the loss function. Different works use a different combination of loss functions; however, most of these choices are only justified empirically. We explore ways to validate these combinations more formally. Overall, our work makes the following contributions:

\begin{itemize}
    \item We separate components of saliency models and discuss the progress on each of them in reference to the literature. Such analysis can help better interpret the models i.e., assess component-wise weaknesses, strengths, and novelty. The analysis allows to optimize saliency models by trying alternates for a particular component while freezing the rest of them. 
    \item We propose an encoder-decoder based saliency detection model called SimpleNet. The main novelty of SimpleNet is a UNet like multi-level integration~\cite{ronneberger2015u}. SimpleNet is fully convolutional; end to end trainable; has lower complexity than counterparts and allows real-time inference. It gives consistent performance over multiple metrics on SALICON and MIT benchmarks, outperforming state of the art models over five different metrics (with significantly notable improvements on KLdiv metric).
    \item We propose a parametric model called MDNSal, which predicts parameters of a GMM instead of a pixel-level saliency map. The main novelty of MDNSal is in readout architecture with a modified Negative Log Likelihood (NLL) loss formulation. It achieves near state of the art performance on SALICON and MIT benchmarks. 
\end{itemize}


\section{Key Components of Saliency Models}

\subsection{Input features}
Early attempts relied on handcrafted low-level features for saliency prediction. Seminal work by Itti~\cite{itti1998model} relied on color, intensity, and orientation maps (obtained using Gabor filter). Valenti~\cite{bylinskii2015saliency} use isophotes (lines connecting points of equal intensity), color gradients, and curvature features. Zhang~\cite{zhang2013saliency} computes saliency maps by analyzing the topological structure of Boolean maps generated through random sampling. Bruce~\cite{bruce2009saliency} use low-level local features (patch level) in combination with information-theoretic ideas. Jude~\cite{judd2009learning} included high-level information by using detectors for faces, people, cars, and horizon. However, most of these methods remain elusive on generic high-level feature representation. 

Recent works are dominated by deep learning architectures owing to their strong performance. Most of this success can be attributed to Convolutional architectures~\cite{kummerer2014deep,kruthiventi2017deepfix,cornia2016deep,jia2018eml}. Some works have also explored combining CNN with recurrent architectures~\cite{cornia2018predicting}. The breakthrough happened via transfer learning of high-level features trained for image classification ~\cite{krizhevsky2012imagenet,simonyan2014very,he2016deep,zoph2018learning,huang2017densely}. The large scale SALICON dataset~\cite{jiang2015salicon} was pivotal in transfer learning process (allowing efficient fine-tuning).  Initial approaches relied on Alexnet or VGG features~\cite{kummerer2014deep,cornia2016deep,liu2018deep}. Other notable architectures like ResNet, DenseNet and NasNet~\cite{cornia2018predicting,jia2018eml} were then explored. Recent works also explore the combinations of features from multiple pre-trained networks~\cite{jia2018eml}. There is enough evidence to agree that using pre-trained features brings significant gains on the task of saliency prediction. We analyze two important design choices: (a) which pre-trained network to pick and (b) should pre-trained weights be frozen or fine-tuned.


\begin{figure*}[t]
\includegraphics[width=\textwidth]{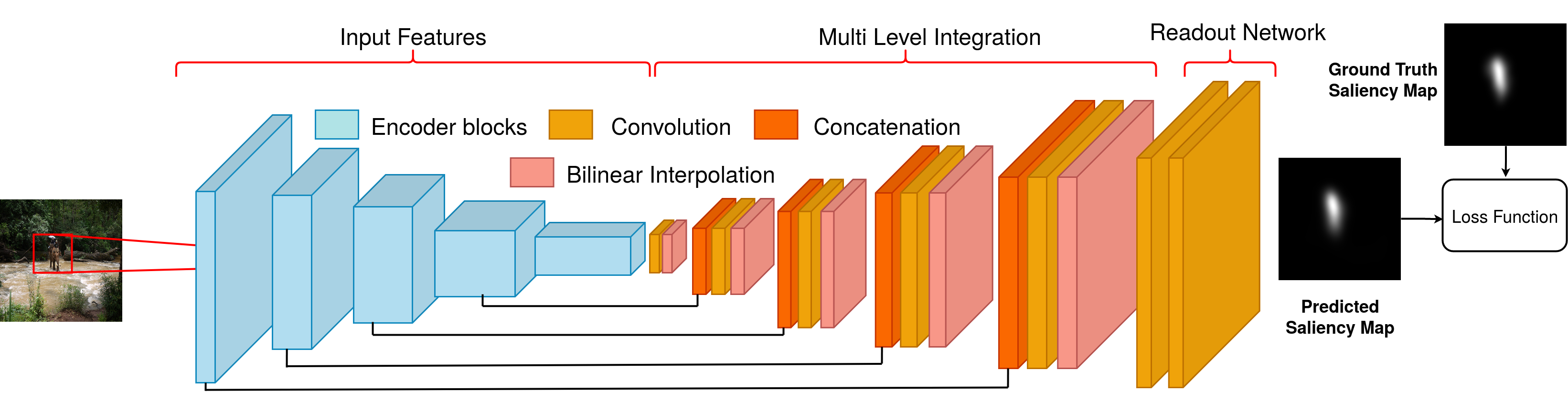}
\caption{SimpleNet Architecture}
\label{fig:SimpleNet}
\end{figure*}

\subsection{Multi-level integration}

It is evident that deep learning models utilizing high-level features significantly outperform the older counterparts, which rely on low-level handcrafted features. However, recent work~\cite{kummerer2017understanding} suggests that the simple low-level model better explains a substantial proportion of fixations when compared to the state-of-the-art model. They quantitatively show this by changing the input features to low-level intensity contrast features (ICF) and keeping the rest of the architecture the same.

Deep networks have employed two main strategies to resolve this concern. The first is to send different image scales as input in parallel. SALICON~\cite{DBLP:conf/iccv/HuangSBZ15} model uses two
different image scales and the idea was then extended to multiple scales~\cite{liu2016learning,li2015visual}. The multi scale spatial stimuli can be tackled by this approach, but not necessarily the semantics. An alternate approach is to take features from different stages of pre-trained CNN. Different levels of semantics (from low, mid, and high-level features) can be thus directly incorporated into the model to resolve the concerns raised in~\cite{kummerer2017understanding}. The work by~\cite{kummerer2014deep} takes a weighted sum of features at different levels post resizing, where the weights are trained through the network. ML-Net~\cite{cornia2016deep} resizes and concatenates features from different levels of VGG-16 model and passes it through additional convolutions layers to predict the map. The work in ~\cite{wang2017deep} individually predicts saliency maps for features from different stages of VGG-19 and then fuses them. In this work, we propose a novel UNet~\cite{ronneberger2015u} like architecture for incorporating multi-level features. The single-stream network with skip connections speeds up training, and the structure allows for an organic hierarchical refinement from high to low-level features (symmetric expansion over high-level context to enable precise localization). 



\subsection{Readout Architectures}
The commonly used readout architecture consists of few convolutional layers post the encoder followed by 1$\times$1 convolutions to control the size of the output saliency map. It is also common to learn an additional prior~\cite{kummerer2014deep,cornia2016deep,kummerer2017understanding}. The prior is often aimed to compensate for the central fixation bias. Work by~\cite{cornia2018predicting} employs an lstm based readout architecture and learns a set of 2D Gaussian priors parameterized by their mean and variance (instead of a single one). In this paper, we employ minimal readout architecture with only convolutional layers. We show that combined with an UNet like multi-resolution encoder; such architecture can outperform state of the art models which rely on priors, complex architectures~\cite{cornia2018predicting} or multi-network feature combinations~\cite{jia2018eml}. 

Interestingly, most of the state of the models directly predict an image as output (the saliency map). Parametric models for computing image saliency have not been explored. Although parametric models come with a bound on the complexity of the model (even if the amount of data is
unbounded), they come with several advantages; especially, they are easier to understand and interpret (Where saliency models should look next?~\cite{bylinskii2016should}). Furthermore, they allow better integration with downstream applications. The importance of predicting distributions has been nicely motivated in~\cite{kummerer2018saliency}. To this end, we propose a novel readout architecture, which directly predicts parameters of a 2D GMM (mean, variance, and mixture weights). The proposed readout architecture can be plugged at the end of any given architecture to output a parametric distribution. We show that, although bounded, the parametric models can achieve near state of the art performance. 

\subsection{Loss functions}

Mean Squared Error (MSE) between predicted and ground truth has been employed as loss function~\cite{kruthiventi2017deepfix}. ML-Net introduced a normalized version of MSE~\cite{cornia2016deep}. 
Most of the recent efforts directly use one of commonly used evaluation metrics or a combination of them as a loss function. The most commonly used loss is computing KL-divergence (KLdiv) between the estimated and ground-truth saliency maps~\cite{DBLP:conf/iccv/HuangSBZ15}. Some papers use a variation of it like negative log likelihood~\cite{kummerer2014deep} or cross entropy~\cite{wang2017deep} instead. Recent works use KLdiv in combination with other metrics like Pearson’s Correlation Coefficient (CC), Normalized Scanpath Saliency(NSS), and Similarity. These combinations bring clear improvement in performance~\cite{jia2018eml,cornia2018predicting}. However, the combinations are often decided empirically. In this work, we provide formal insights to choose a minimal and comprehensive loss function.

\section{Proposed Architectures}

\begin{figure*}[t]
\includegraphics[width=\textwidth]{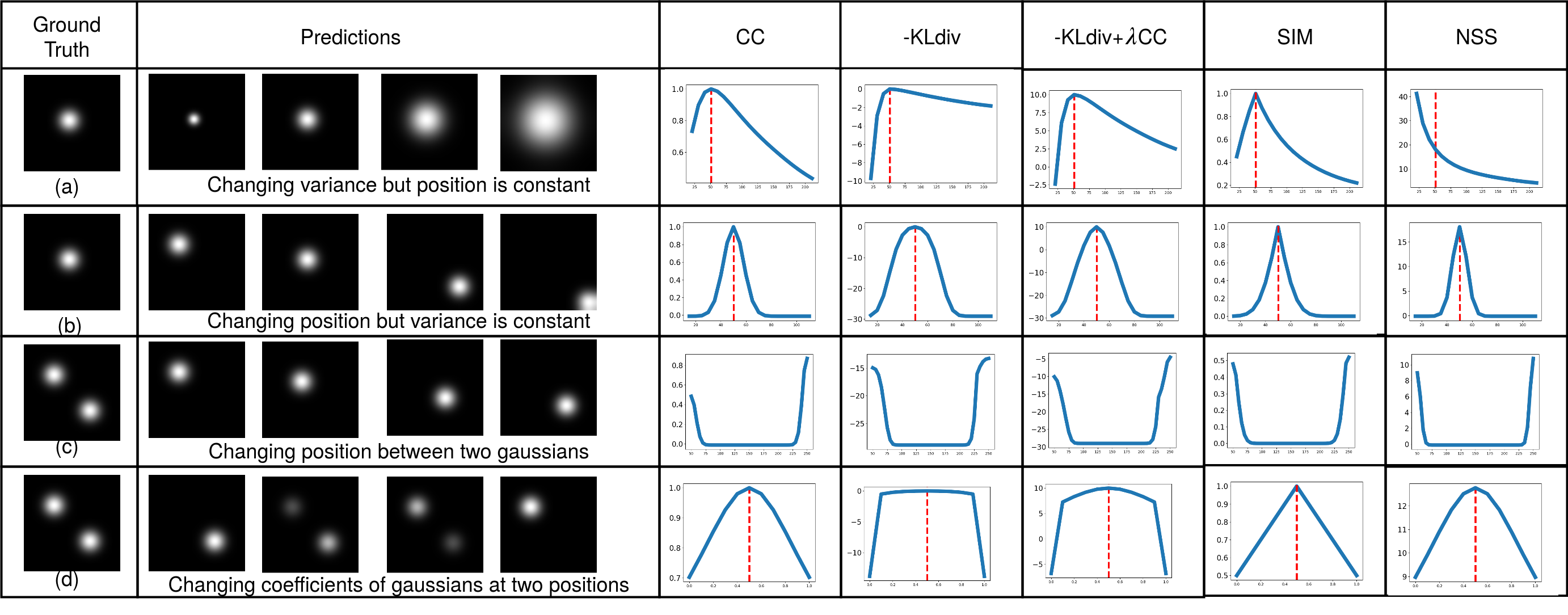}
\caption{We synthetically varied saliency predictions w.r.t the ground truth in order to quantify effects on the loss functions. Each row corresponds to varying a single parameter value of the prediction: (a) Variance, (b) location on a single mode, (c) location between two modes, (d) relative weights between two modes. The x-axis spans the parameter range and the dotted red line corresponds to the ground truth (if applicable).}
\label{fig:loss}
\end{figure*}

\begin{figure*}[t]
\includegraphics[width=\textwidth]{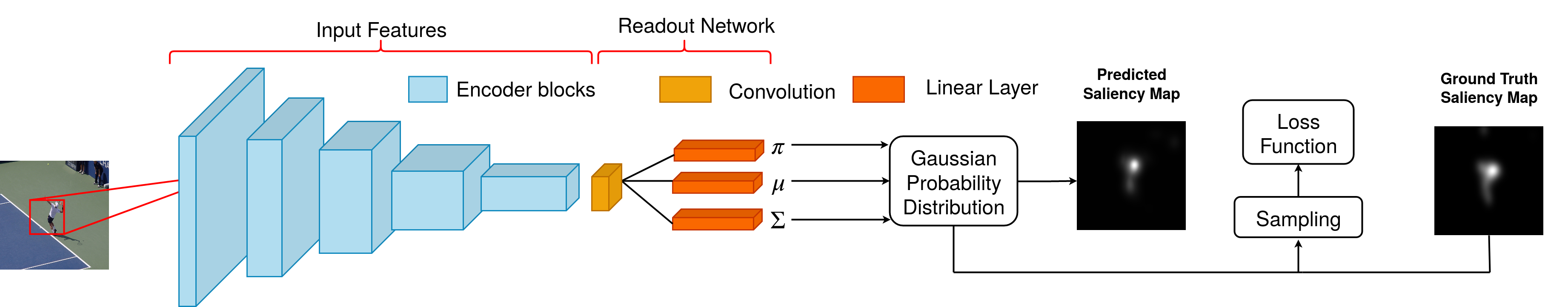}
\caption{MDNSal Architecture}
\label{fig:MDNSal}
\end{figure*}

We propose two end-to-end architectures SimpleNet and MDNSal. SimpleNet is an encoder-decoder architecture that predicts the pixel-wise saliency values, while MDNSal is a parametric model that predicts parameters of a GMM distribution. We now describe each of the models in detail.


\subsection{SimpleNet}

The overall architecture of the SimpleNet is shown in Figure~\ref{fig:SimpleNet}. It is a fully convolutional, single-stream encoder-decoder architecture, which is end to end trainable. The name SimpleNet is derived from the design where the goal was to keep each component simple and minimal, without compromising the performance. 

\paragraph{Input features} SimpleNet directly takes input from the pre-trained architectures designed for image classification. We explore four different architectures VGG-16, ResNet-50, DenseNet-161, and PNASNet-5 and compare their performance. The feature extraction layers (shown as encoder blocks in Figure~\ref{fig:SimpleNet}) are initialized with the pre-trained weights and later fine-tuned for the saliency prediction task. 

\paragraph{Multi-level integration} SimpleNet employs a UNet like architecture that symmetrically expands the input features starting at the final layer of the encoder (input features). The symmetric expansion enables precise localization. Every step of the expansion consists of an upsampling of the feature map, a concatenation with the corresponding scale feature map from the encoder. The number of channels are then reduced using 3$\times$3 convolutions followed by ReLU. 

\paragraph{Readout architecture} The readout architecture consists of two 3$\times$3 convolutional layers; the first is followed by ReLU, and the second uses a sigmoid to output the saliency map.

\paragraph{Loss function} The loss function compares the output saliency map with the ground truth. We use a combination of Kullback-Leibler Divergence(KLdiv) and Pearson Cross Correlation (CC) metrics as a loss function. KLdiv is an information-theoretic measure of the difference between two probability distributions:
\begin{equation}
    KLdiv(P,Q) = \sum_{i} Q_{i}\log(\epsilon + \frac{Q_{i}}{P_{i} + \epsilon} ),
\end{equation}
where $P$, $Q$ are predicted and ground truth maps respectively and $\epsilon$ is a regularization term.\\
CC is a statistical method used generally in the sciences for measuring how correlated or dependent two variables are
\begin{equation}
    CC(P,Q) = \frac{\sigma(P,Q)} {\sigma(P) \times 
    \sigma(Q)}. 
\end{equation}

\begin{table*}[t]
    \caption{Validation results on all three studied datasets}
    \label{tab:val_simple_net}
    \centering
     \footnotesize
    \begin{tabular}{|l|rrrrr|rrrrr|}
    \hline
     & \multicolumn{5}{c|}{SimpleNet} &  \multicolumn{5}{c|}{MDNSal} \\
        & KLdiv  & CC     & AUC    & NSS    & SIM & KLdiv  & CC     & AUC    & NSS    & SIM\\
    \hline
    \textbf{SALICON}                  & 0.193 & 0.907 & 0.871  & 1.926 & 0.797 & 0.217 & 0.899 & 0.868 & 1.893 & 0.797\\
    \textbf{MIT1003}                  & 0.558 & 0.786 & 0.907 & 2.870 & 0.626& 0.634 & 0.779 & 0.904 & 2.814 &0.624 \\
    \textbf{CAT2000}                  & 0.256  & 0.895 & 0.883 & 2.400 & 0.758& 0.293 & 0.889 & 0.878 & 2.329 &0.751 \\
    \hline
    \end{tabular}
   
\end{table*}

\begin{table}[t]
     \caption{SimpleNet's validation results on SALICON with various Encoders}
    \label{tab:val_simple_net_enc}
    \centering
    \footnotesize
    \begin{tabular}{|l|rrrrr|}
    \hline
        \textbf{Models} & CC & NSS & KLdiv & AUC & SIM\\
    \hline
    \textbf{VGG-16}                   & 0.871  & 1.863 & 0.238 & 0.864 & 0.772\\
    \textbf{ResNet-50}                  & 0.895 & 1.881 & 0.211 & 0.868 & 0.786\\
    \textbf{DenseNet-161}                  & 0.902 & \textbf{1.930} & 0.210 & 0.87 &0.795 \\
    \textbf{PNASNet-5}                  & \textbf{0.907} & 1.926 & \textbf{0.193} & \textbf{0.871} & \textbf{0.797} \\
    \hline
    \end{tabular}
\end{table}

\begin{table}[t]
    \caption{SimpleNet's validation results on SALICON with various Loss Functions}
    \label{tab:val_simple_net_loss}
    \centering
     \footnotesize
    \begin{tabular}{|l|rrrrr|}
    \hline
       \textbf{Loss Functions} & CC & KLdiv &  NSS & AUC & SIM \\
    \hline
    \textbf{KL}                 & 0.904 & 0.223 & 1.935 & 0.870 & 0.797 \\
    \textbf{KL + CC}            & \textbf{0.906} & \textbf{0.192} & 1.925 & 0.871 & \textbf{0.798} \\
    \textbf{KL + CC + NSS}      & 0.900 & 0.204 & \textbf{1.998} & \textbf{0.872} & 0.794 \\
    \hline
    \end{tabular}
\end{table}

\begin{table}[t]
     \caption{MDNSal's validation results on SALICON with various number of Gaussians}
    \label{tab:val_mdn_c}
    \centering
     \footnotesize
    \begin{tabular}{|l|rrrrr|}
    \hline
       \textbf{No of Gaussians($C$)} & CC & KLdiv &  NSS & AUC & SIM \\
    \hline
    \textbf{8}                 & 0.882 & 0.256 & 1.849 & 0.864 & 0.778 \\
    \textbf{16}            & 0.892 & 0.240 & 1.881 & 0.867 & 0.787 \\
    \textbf{24}      & 0.895 & 0.233 & 1.887 & 0.867 & 0.789 \\
    \textbf{32}      & \textbf{0.899} & \textbf{0.224} & \textbf{1.892} & \textbf{0.868} & \textbf{0.797} \\
    \textbf{48}      & 0.896 & 0.231 & 1.889 & \textbf{0.868} & 0.790 \\
    \textbf{64}      & 0.896 & 0.230 & \textbf{1.892} & \textbf{0.868} & 0.790 \\
    \hline
    \end{tabular}
\end{table}
 The combination of KLdiv and CC is motivated by the analysis presented in Figure~\ref{fig:loss} (the analysis is inspired by ~\cite{bylinskii2018different}). KLdiv is highly sensitive to false negatives and results in steeper costs (consider case (a), (b) in Figure~\ref{fig:loss}). Steeper costs lead to larger gradients, which are crucial in initial training (which motivates the use of KLdiv or its variants as a backbone of loss function). CC is symmetric in terms of false positives and false negatives and gives typical behavior in each scenario of Figure~\ref{fig:loss}. The combination provides appropriate behavior in each of the studied scenarios while maintaining steeper costs (scenarios (a), (d) (Figure~\ref{fig:loss}). We also explore the use of Normalized Scanpath Saliency (NSS) as a loss function in the ablation studies.

\subsection{MDNSal}
The overall architecture of the MDNSal is shown in Figure~\ref{fig:MDNSal}. The network is inspired by the literature on Mixture Density Networks~\cite{bishop1994mixture}. 

\paragraph{Input features} Similar to SimpleNet; we apply transfer learning from pre-trained image classification networks. The fine-tuning of the pre-trained features is extremely crucial in MDNSal and leads to significant performance improvements.  

\paragraph{Multi-level integration} MDNSal only uses the features from the last convolutional layer of the pre-trained networks and is devoid of multi-level integration. Since the outputs are parameters instead of a per-pixel map, the multi-level features do not play a significant role.  

\paragraph{Readout architecture} The readout architecture consists of a convolutional layer to reduce the number of channels followed by a ReLU. The output is then passed to three parallel fully connected layers predicting mixture weight ($\pi$), mean ($\mu$), and the covariance matrix ($\Sigma$) for each Gaussian. For $C$ mixtures, the sizes of the output layers are $C$, $C\times2$ and $C\times2$ for $\pi$, $\mu$ and $\Sigma$ respectively. We predict only the diagonal elements of the covariance matrix $\Sigma$. 

\paragraph{Loss function}
We define Negative log-likelihood (NLL) loss function to train the parameters of the Gaussians as follows:



 
 \begin{equation}
    NLL(P,Q) = -\sum_{i}q_{i} \log(p_{i} + \epsilon ).
\end{equation}

Where $i$ represents exhaustive sampling across spatial dimensions of the image, $p_i$ is the likelihood of the sampled point to fit the distribution with C Gaussians, $q_i$ is the corresponding ground truth value and $\epsilon$ is a small constant. $p_i$ is further defined as follows:

 \begin{equation}
     p(x; \pi, \mu, \Sigma) = \sum_{c=1}^{C} \pi_{c} \frac{1}{\sqrt{(2\pi_{c})\textsuperscript{2} |\Sigma_{c}|}} e^{-\frac{1}{2}(x-\mu_{c})^T\Sigma_{c}\textsuperscript{-1}(x-\mu_{c})}.
 \end{equation}
 
Similar to SimpleNet we use a combination of NLL and CC for training MDNSal.

\section{Experiments and Results}
\subsection{Datasets}
\paragraph{SALICON:} We use SALICON dataset for training our models. It consists of 10,000 training, 5000 validation, and 5000 test images. It is the largest dataset for saliency prediction and was labeled based on mouse-tracking (shown to be equivalent to the eye-fixations recorded with an eye-tracker).  Our experiments are based on the newest release, SALICON 2017, from the LSUN challenge.
\paragraph{MIT300:} MIT300 test dataset consists of 300 natural images with eye-tracking data of 39 observers. The labels of MIT300 are non-public. We use MIT1003 consisting of 1003 images to fine-tune the initial model trained on SALICON dataset. We have done 10 fold cross-validation by splitting the images into 903 train, and 100 validation and have chosen the best model for test submission. 
\paragraph{CAT2000: } CAT2000 consists of 4000 images(2000 train and 2000 test) taken from 20 different categories like Action, Art, Cartoon etc., with eye-tracking data from 24 observers. Similar to MIT1003, we split the images into 1800 train and 200 validation and perform fine-tuning on our model. 

\begin{table*}[t]
    \caption{Our proposed models performance on the SALICON TEST.}
    \label{tab:test_salicon}
    \centering
     \footnotesize
    \begin{tabular}{|l|rrrrrrr|} 
    \hline
                     & KLdiv$\downarrow$          & CC$\uparrow$             & AUC$\uparrow$            & NSS$\uparrow$           & SIM$\uparrow$            & IG$\uparrow$            & sAUC$\uparrow$           \\
    \hline
    EMLNET~\cite{jia2018eml}           & 0.520           & 0.886          & \textcolor{blue}{0.866}          & \textcolor{red}{2.050} & 0.780           & 0.736         & \textcolor{red}{0.746} \\
    SAM-Resnet~\cite{cornia2018predicting}  & 0.610           & \textcolor{blue}{0.899}          & 0.865          & \textcolor{blue}{1.990}          & \textcolor{red}{0.793}          & 0.538         & 0.741          \\
    MSI-Net~\cite{alex2019contextual}       & 0.307          & 0.889          & 0.865          & 1.931         & 0.784          & 0.793         & 0.736          \\
    GazeGAN~\cite{che2019gaze}              & 0.376          & 0.879          & 0.864          & 1.899         & 0.773          & 0.720          & 0.736          \\
    MDNSal (Ours)                          & \textcolor{blue}{0.221}          & \textcolor{blue}{0.899}          & 0.865          & 1.935         & \textcolor{blue}{0.790}           & \textcolor{blue}{0.863}         & 0.736          \\ 
    SimpleNet (Ours)                        & \textcolor{red}{0.201} & \textcolor{red}{0.907} & \textcolor{red}{0.869} & 1.960          & \textcolor{red}{0.793} & \textcolor{red}{0.880} & \textcolor{blue}{0.743}          \\
    \hline
    \end{tabular}
\end{table*}

\begin{table*}[]
    \caption{Our proposed models performance on the MIT Saliency Benchmark.}
    \label{tab:test_mit}
    \centering
     \footnotesize
    \begin{tabular}{|l|rrrrrrr| }
    \hline
                                                & KLdiv$\downarrow$         & CC$\uparrow$                            & AUC$\uparrow$                           & NSS$\uparrow$                           & SIM$\uparrow$           & sAUC$\uparrow$          & EMD$\downarrow$           \\
    \hline
    EMLNET~\cite{jia2018eml}                    & 0.84          & \textcolor{blue}{0.79}         & \textcolor{red}{0.88}          &\textcolor{red}{2.47}           & \textcolor{blue}{0.68}          & 0.70          & \textcolor{red}{1.84}   \\
    DeepGaze2~\cite{kummerer2017understanding}        & 0.96          & 0.52                          & \textcolor{red}{0.88}          & 1.29                         & 0.46           & \textcolor{blue}{0.72}         & 3.98          \\
    SALICON~\cite{DBLP:conf/iccv/HuangSBZ15}    & 0.54          & 0.74                          & \textcolor{blue}{0.87}         & 2.12                         & 0.60          & \textcolor{red}{0.74} & 2.62          \\
    DPNSal~\cite{oyama2018influence}            & 0.91          & \textcolor{red}{0.82}         & \textcolor{blue}{0.87}         & \textcolor{blue}{2.41}          & \textcolor{red}{0.69} & \textcolor{red}{0.74} & 2.05          \\
    DenseSal~\cite{oyama2017fully}              & 0.48          & \textcolor{blue}{0.79}                          & \textcolor{blue}{0.87}        & 2.25                          & 0.67          &\textcolor{blue}{ 0.72}          & 1.99          \\
    DVA~\cite{wang2017deep}                     & 0.64          & 0.68                          & 0.85                          & 1.98                          & 0.58          & 0.71          & 3.06          \\ 
    MDNSal (Ours)             & \textcolor{blue}{0.47}         & 0.78                          & 0.86                          & 2.25                          & 0.67          & 0.71          & \textcolor{blue}{1.96}          \\
    SimpleNet (Ours)           & \textcolor{red}{0.42}          & \textcolor{blue}{0.79}        & \textcolor{blue}{0.87}          & 2.30                         & 0.67          & 0.71          &  2.06         \\
    \hline
    \end{tabular}
\end{table*}

\subsection{Experimental Setup}
We resize the input images into 256x256 resolution for both the models. We train SimpleNet for 10 epochs with learning rate starting from 1e-4 and reducing it after 5 epochs. MDNSal is trained for 50 epochs with learning rate 1e-4. We use 32 Gaussians ($C=32$) in MDNSal. Backpropagation was performed using ADAM optimizer in both the networks. The model trained on SALICON was fine-tuned using MIT1003 and CAT2000. We submitted the test results for SALICON to LSUN17\footnote{http://salicon.net/challenge-2017/} and MIT300 test results to MIT Saliency Benchmark\footnote{http://saliency.mit.edu}. We only present validation results on the CAT2000 dataset.



\subsection{Ablation Analysis}
We examine the effects of (a) changing the input feature, (b) using different combinations of the loss function, and (c) the significance of hierarchy. The analysis is made using SimpleNet model on SALICON validation set. Table~\ref{tab:val_simple_net_enc} illustrates the results by varying the pre-trained network for the input feature component. PNASNet-5 achieves the best overall results and is used as the backbone for all the following experiments. Ablation results by adding CC and NSS to the KLdiv loss are presented in Table~\ref{tab:val_simple_net_loss}. Adding CC improves the performance over just using KLdiv loss (on both KL and CC metrics). Higher performance on the NSS metric can be achieved by adding an NSS term to the loss; however, it brings minor reductions in the KLdiv and CC metric. To keep things minimal, we use KLdiv+CC loss for later experiments. 

We also explore the significance of multi-level integration by learning SimpleNet by just using the last conv layer of PNASNet-5. The CC drops to 0.89 from 0.907, and KLdiv increases to 0.22 from 0.193, indicating the importance of multi-level integration. Finally, the results of the validation set on all three datasets are presented in Table~\ref{tab:val_simple_net}.

We perform another set of experiments on MDNSal. The first ablation experiment is aimed to understand the impact of  changing the number of Gaussians($C$). The results are presented in Table~\ref{tab:val_mdn_c} and it is observed that using 32 Gaussians gives best performance and thus we use the same number of mixtures in the later experiments. As next experiment, we relax the constraint of only predicting the diagonal values of the covariance matrix. We predict the full covariance matrix with positive-definite constraints, which is necessary to compute the loss. To enforce this constraint we adopt the method by ~\cite{dorta2018structured}, where we predict lower triangular matrix($L$) and get covariance matrix using $A = LL^{T}$ and $\Sigma^{-1} = A$. Using full covariance matrix however did not give any visible improvements (CC remained same to 0.899 on SALICON validation set). Hence, we use the diaognal approximation in all the remaining experiments on MDNSal. 






\subsection{Comparison with state of the art}
We quantitatively compare our models with state of the art on SALICON and MIT300 test sets. Table~\ref{tab:test_salicon} shows the results on the SALICON dataset in terms of KLdiv, CC, AUC, NSS, SIM, IG, and sAUC metrics. SimpleNet gives consistent results on all seven metrics. It achieves the best performance on five different metrics and outperforms state of the art by a large margin on KLdiv and IG. We are third-best in NSS metric; however, if crucial, that can be compensated by adding an NSS loss, as indicated in the ablation studies. Although parametric, surprisingly, MDNSal also gives competent performance across various metrics (only second to SimpleNet on four metrics). SimpleNet and MDNSal also achieve state of the art performance on MIT300 test dataset, as shown in Table~\ref{tab:test_mit}. SimpleNet gives the best results on KLdiv and is competent in all other metrics. MDNSal gives a similar performance with the second-best results on KLdiv and EMD.

The work by~\cite{bylinskii2018different} recommends CC as one of the ideal metrics to report, as it makes limited assumptions about input format and treats both false positives and negatives symmetrically. They further suggest KL and IG as good choices concerning benchmark intended to evaluate saliency maps as probability distributions. Both our models give a leading performance on KLdiv and CC on both MIT300 and SALICON test sets, which makes them an ideal choice for the task of saliency prediction. 

\begin{figure*}[h]
\includegraphics[width=\textwidth]{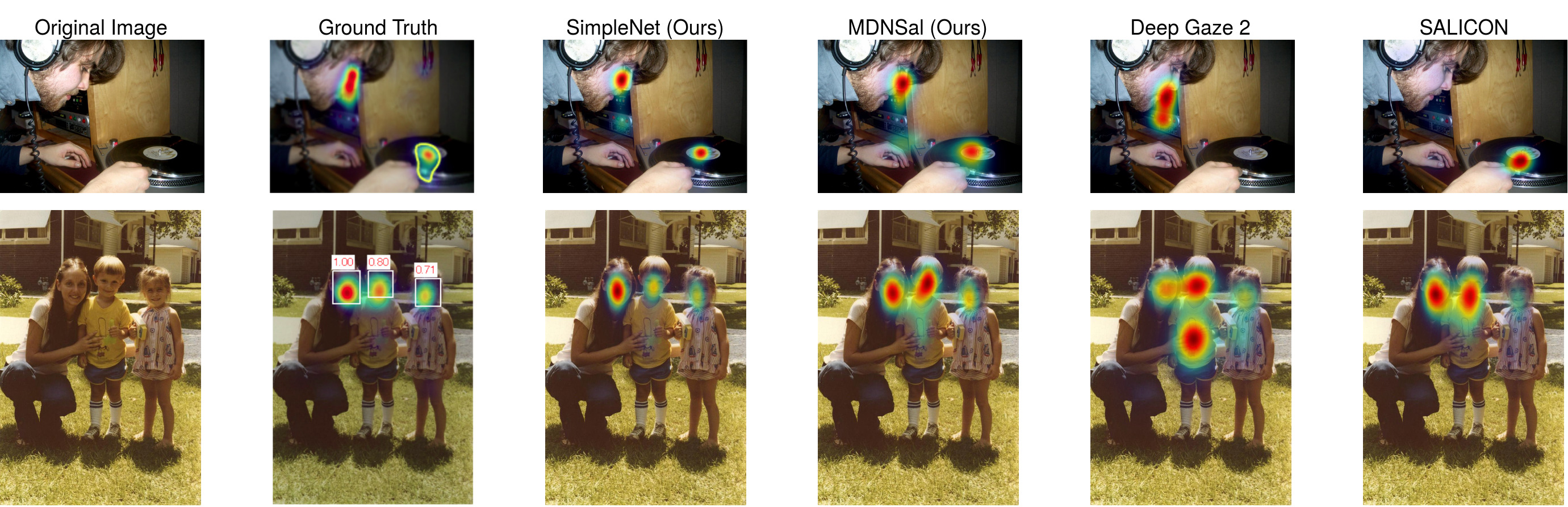}
\caption{Examples of predicted saliency maps. Both images are taken from MIT300 test set and ground truth images are taken from~\cite{bylinskii2016should}. We compare results of the proposed SimpleNet and MDNSal models with other state of the art approaches. First row (action): SimpleNet and MDNSal accurately predicts both the person's face and where he is looking (indicated by yellow boundary). In contrast other models miss out on either the action or the face. Second row (Faces with relative importance): our model gives accurate localization on all three faces and preserves relative importance. }
\label{fig:motivation1}
\end{figure*}

We qualitatively compare results of the proposed models with other state of the art methods. The results on couple of images from MIT300 test dataset are shown in Figure~\ref{fig:motivation1}. The ground truth images are chosen from the carefully curated set in~\cite{bylinskii2016should}. Our model performs well both in terms on coverage (predicting all the salient regions), accurate localization and the relative order of importance. 




\section{Conclusion}
In this paper, we identify four key components of the saliency detection architectures and analyze how the previous literature has approached the individual components. The analysis helps to explore agreements, redundancies, gaps, or need for optimization over these components. Using that as a basis, we propose two novel architectures called SimpleNet and MDNSal. SimpleNet improves upon the encoder-decoder architectures, and MDNSal opens up a new paradigm of parametric modeling. Both models are devoid of complexities like prior maps, multiple input streams, or recurrent units and still achieve the state of the art performance on public saliency benchmarks. Our work suggests that the way forward is not necessarily to design more complex architectures but a modular analysis to optimize each component and possibly explore novel (and simpler) alternatives.

\bibliographystyle{IEEEtran}
\bibliography{IEEEabrv,IEEEexample}

\end{document}